\documentclass[11pt, a4paper, twocolumn]{article}

\usepackage{color}
\usepackage[utf8]{inputenc}
\usepackage[english]{babel}
\usepackage[top=2cm, bottom=2cm, left=1.9cm, right=1.9cm]{geometry}

\usepackage{titlesec}
\titlespacing{\section}{0pt}{\parskip}{\parskip}
\titlespacing{\subsection}{0pt}{\parskip}{\parskip}
\titlespacing{\subsubsection}{0pt}{\parskip}{\parskip}

\usepackage{indentfirst}
\usepackage[export]{adjustbox}
\usepackage{graphicx}
\usepackage{amssymb}
\usepackage{amsmath}

\usepackage[dvipsnames]{xcolor}

\usepackage{caption}
\usepackage[figurename=Fig.]{caption}
\usepackage[colorlinks, citecolor=ForestGreen]{hyperref}

\usepackage{cite}

\usepackage{abstract}
\usepackage{multirow}
\usepackage{array}
\newcolumntype{*}{>{\global\let\currentrowstyle\relax}}
\newcolumntype{^}{>{\currentrowstyle}}

\title{\textbf{A Hierarchical Approach to Remote Sensing Scene Classification}}
\date{\vspace*{2pt}}

\usepackage{authblk}
\author[1]{Ozlem Sen\thanks{Corresponding author}}
\author[1, 2]{Hacer Yalim Keles}

\affil[1]{\footnotesize Ankara University, Computer Engineering Department, Turkey}
\affil[2]{\footnotesize Hacettepe University, Computer Engineering Department, Turkey}
\affil[ ]{\footnotesize ozlem.sen@gmail.com, hacerkeles@cs.hacettepe.edu.tr}

\usepackage{ulem}

\begin{document}
\maketitle

\begin{abstract}{
		Remote sensing scene classification deals with the problem of classifying land use/cover of a region from images. To predict the development and socioeconomic structures of cities, the status of land use in regions is tracked by the national mapping agencies of countries. Many of these agencies use land-use types that are arranged in multiple levels. In this paper, we examined the efficiency of a hierarchically designed Convolutional Neural Network (CNN) based framework that is suitable for such arrangements. We use the NWPU-RESISC45 dataset for our experiments and arranged this data set in a two-level nested hierarchy. Each node in the designed hierarchy is trained using DenseNet-121 architectures. We provide detailed empirical analysis to compare the performances of this hierarchical scheme and its non-hierarchical counterpart, together with the individual model performances. We also evaluated the performance of the hierarchical structure statistically to validate the presented empirical results. The results of our experiments show that although individual classifiers for different sub-categories in the hierarchical scheme perform considerably well, the accumulation of the classification errors in the cascaded structure prevents its classification performance from exceeding that of the non-hierarchical deep model.  
}\end{abstract}


\section{Introduction}

Thanks to the available satellite and aerial imaging technologies, the earth's surface can be observed and analyzed with images in various spatial, spectral, and temporal resolutions. The research in remote sensing scene image analysis aims to classify scene images that are obtained from available instruments in various forms to some predefined, discrete set of semantic categories, e.g. residential area, forest, bridge, etc. In this context, a scene image refers to a manually created local image patch that ideally contains a scene belonging to a particular semantic category obtained from large-scale remote sensing images \cite{Ref-NW45}. Automating the content labeling of local patches in large scale remote sensing images is useful for many application areas such as environment monitoring \cite{phinn2012multi}, urban planning \cite{
li2013object,mishra2014mapping,kim2009forest}, analysis of agricultural regions \cite{janssen1992knowledge}, managing natural disasters \cite{stumpf2011object,cheng2013automatic}, land use/cover determination \cite{chen2014pyramid,chen2016land} and so on \cite{Cheng_2020}.

Early works in this domain focused more on the identification of pixel labels in aerial scenes since the resolution of such images are very low and sometimes, a sub-pixel analysis may also be necessary \cite{janssen1992knowledge}. However, the emergence of high-resolution images made interpretation of pixels harder and meaningless, since the semantic context of a scene can be interpreted by considering the relationships of multiple pixels, belonging to different objects and their surroundings \cite{blaschke2001s}. The next direction was the identification of objects, such as buildings, cars, trees, etc., from remote sensing images \cite{blaschke2014geographic,druaguct2006automated,eisank2011generic}. However, in higher resolution images, semantic content is identified correctly with the interaction of multiple objects, or some of the content is ambiguous without the surrounding context if only object appearances are taken as references; most object-classification based approaches struggle among similar scenes in the scene classification task. For instance, trees, roads, and houses could be visible in a dense residential area. The content of the high-resolution remote sense images are usually heterogeneous; hence, closing the semantic gap between the low-level semantics of the image pixels/objects and high-level (i.e. scene level) semantics become crucial \cite{bratasanu2010bridging}. Therefore, recent works focus more on a semantic-level remote sensing scene classification \cite{zhao2016feature, xia2015accurate, cheng2015effective,zhang2013semi, zheng2012automatic,Ref-UCM,zhong2015scene,zhang2016semantic,hu2015unsupervised,Ref-NW45}.

Since the publication of the UC Merced dataset in 2010 \cite{Ref-UCM}, various remote sensing scene image datasets have been made publicly available to support the research in the remote sensing scene classification (RSSC) domain. Most of the related publicly available datasets provide images in red, green and blue (RGB) spectral bands, in various spatial and image resolutions \cite{Ref-UCM, Ref-AID, Ref-WHU, Ref-RSICB, Ref-NW45}. Among these, the most challenging dataset, which is NWPU-RESIS45 (NW45 for short),  provides more samples per 45 distinct semantic scene categories with high image variations and diversity, in a range of spatial resolution from 0.2 to 30 meters. In this dataset, scenes contain a broad category of images including land use and land cover classes (e.g., farmland, forest, residential area), man-made object classes (e.g., airplane, ship, bridge, church), and natural object classes (e.g., beach, river, cloud, etc.).

Remote sensing scene classification (RSSC) is a challenging problem, mainly due to high variations in pose and appearance, viewpoint, illumination, and backgrounds of the scene images. The images have high intra-class variation and inter-class similarities. There are various computational approaches for the automatic classification of remote sensing scene images in the literature (Section \ref{sec:2}). Recent advancements in the computer vision and machine learning domains, particularly deep learning approaches, accelerated the success of the computational approaches considerably in this field \cite{Cheng_2020}. The most successful performances are achieved using the convolutional neural network (CNN) based models in this domain. Promising classification accuracies are obtained by applying transfer learning to some pre-trained deep CNN models that are trained on the ImageNet dataset \cite{deng2009imagenet}. Adaptation of the models to the RSSC domain improves classification performances, which is otherwise low due to the relative scarcity of the number of images that are required for training deep models in the available RSSC benchmarks \cite{Ref-UCM, Ref-AID, Ref-WHU, Ref-RSICB, Ref-NW45}.

Most of the existing solutions to the RSSC problem use deep CNN-based models to directly predict the fine-grained class labels from a given scene image. In this work, we investigate the RSSC problem from a very basic, yet unexplored perspective; using a two-layer hierarchical structure, where we utilize the semantic relationships between the classes to define the hierarchy. Given the hierarchy tree, we construct a hierarchical model to solve the problem by using high capacity deep CNN models in each node of the hierarchy tree. Although some hierarchic solutions exist in the literature in the RSSC domain, the hierarchy is usually defined based on clustering similarities of the collected samples' features.

 \begin{figure}[t!]
	
	\centering
	\includegraphics[width=0.5\textwidth]{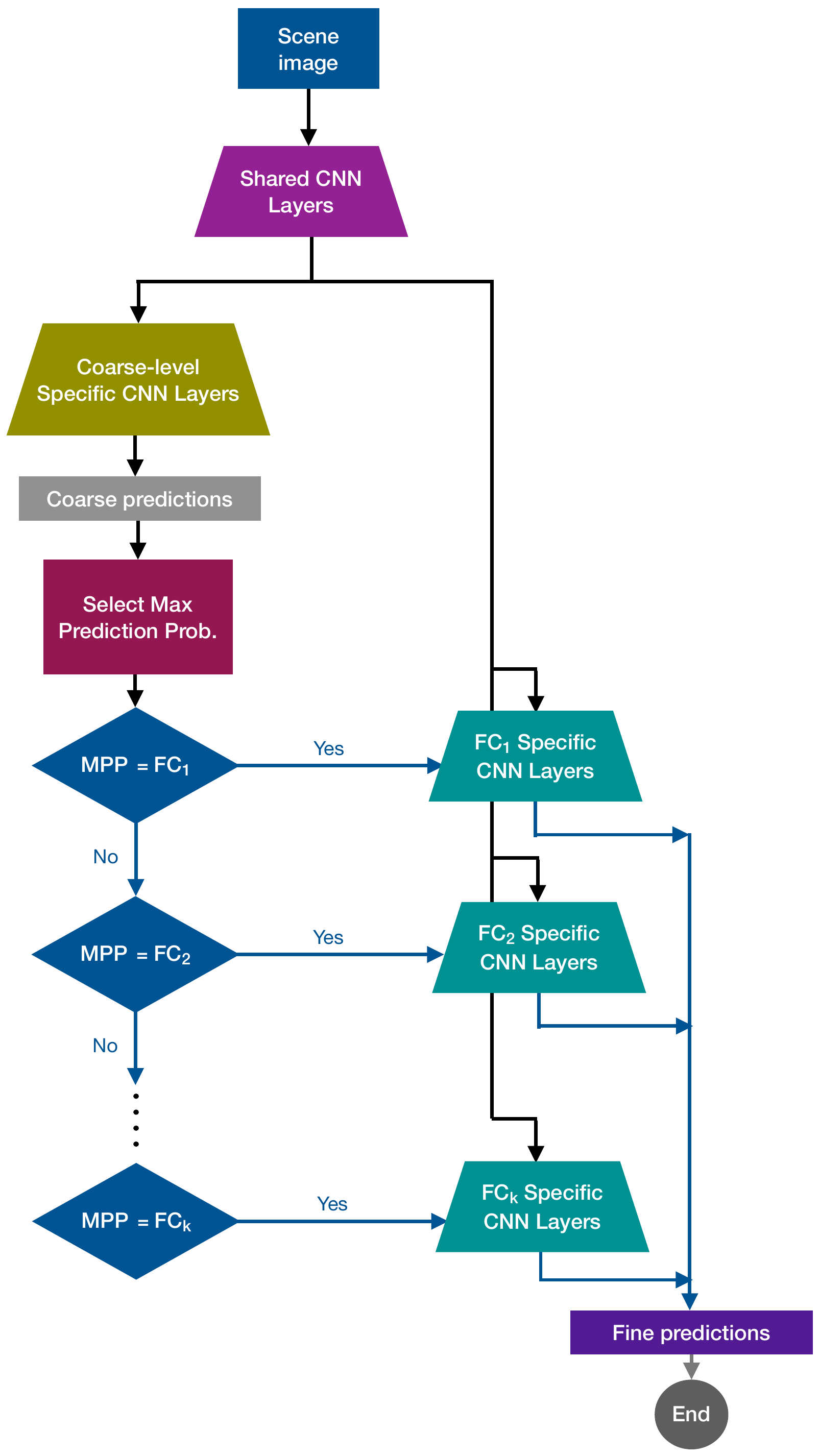}
	\caption{Flow diagram of the proposed solution.} 
	\label{fig:Fig_Flow}
\end{figure}

Our hierarchic solution has two primary advantages: (1) More general distinguishing features of a category that are ignored by a fine-level category label can be captured in a coarse-level classifier, (2) a dedicated classifier for hard-to-distinguish classes can learn more discriminative features utilizing a larger dedicated learning capacity. With this in mind, we designed a two-level hierarchical framework utilizing the defined semantic categories of the scene classes, \textit{i.e. buildings, transportation, water areas, natural lands, and constructed lands}, then each node in this representation is sub-divided into related fine-grained classes, \textit{(e.g. beach, lake, river, etc. under the water areas category)}. We have a backbone CNN (pre-trained DenseNet  \cite{huang2018densely}) that serves as the building block in our solution; we then fine-tune the coarse-level independent layers in the first layer of our hierarchy and the fine-level independent layers in the second layer of the hierarchy. The models are trained at each node independently, sharing the low-level features of the building block CNN. The flow diagram of our model is depicted in Fig. \ref{fig:Fig_Flow}.

Our model is similar more to the hierarchical model proposed in HD-CNN architecture \cite{ICCV2015_HDCNN}, which is designed for object classification tasks for CIFAR100 and ImageNet benchmark datasets; yet differently from them, we conditionally run the fine category model components considering the coarse-level maximum prediction probabilities. There are two primary reasons behind our design: (1) We want to be able to evaluate the independent performances of the fine-level classifiers, in case it is preferred in particular domains, (2) to optimize the computation time of the hierarchic model, by truncating alternative paths in the hierarchy tree considering the likelihoods. We reduce the computational time considerably by the conditional execution of the fine-level CNN models without sacrificing the model performance; only one branch, with the maximum prediction probability, is executed for each image.

In this research, we want to inquire about the effects of the increased capacity utilization on model performance in the RSSC domain, particularly for addressing the ambiguities arising from high intra-class variances and inter-class similarities in scenes. Even though existing deep learning models have high capacities, i.e. capable of learning and using millions of parameters for effective non-linear transformations, the nature of the data in the RSSC domain is still challenging due to the ambiguities in the class boundaries \cite{sen2019scene}. In this context, we experimented with a hierarchical scheme that we designed in a two-layer tree structure, where each node of the tree is itself a deep model dedicated to classifying a subset of semantically grouped fine-grained categories. We worked with (NW45) \cite{Ref-NW45} by re-arranging the categories in this dataset in two layers. The contributions of this paper can be summarized as follows:

\begin{itemize}
\item We compared the performances of a hierarchical (two-level) and a non-hierarchical (one-level) models using the class hierarchies that we formed from the NW45 dataset. Both schemes are designed using the same backbone CNN model architectures, i.e. DenseNet-121. To the best of our knowledge, this is the first time in this domain that the semantic hierarchy is created as we presented in this work, using independent deep CNN models at the nodes of the hierarchy tree.
 \item We computed the approximate overall prediction probabilities of the hierarchical model branches statistically by taking account of the independent prediction performances of the individual models in the structure. The empirical results are validated with these estimates. 
\item Even though we applied pruning in the execution of hierarchical tree branches, by following only the most likely branch in the second level for computational efficiency, the performance of the hierarchical model is comparable to the state-of-the-art non-hierarchic model performances. 
\item  We empirically observed and statistically validated that due to the accumulation of errors in the cascading layers of the hierarchical structure, the overall performance of the hierarchical model can not exceed that of the similarly designed non-hierarchical model. Yet, the independent fine-grained classifier accuracies in the second level of our hierarchical structure are higher as we expected and can be utilized in related contexts as separate models.  
\end{itemize}

 The remainder of this paper is organized as follows: In Section \ref{sec:2}, we provide related works particularly focusing on deep learning-based approaches in the RSSC domain. In Section \ref{sec:3}, we explain our dataset and CNN models. Following that, in Section \ref{sec:32}, we provide our proposed method in detail. In Section \ref{sec:4}, we provide our experiment results, and finally, in Section \ref{sec:5}, we provide the conclusion part.
 

\section{Related Works} \label{sec:2}
From a broad perspective, three main classification approaches exist in the RSSC literature \cite{Cheng_2020}: pixel-level, object-level and scene-level approaches. Pixel-level methods are based on semantic segmentation, where each pixel in an image is assigned to one of the pre-defined semantic categories. It is an active research topic, particularly for the multispectral and hyperspectral analysis of the scene images \cite{tuia2011survey, he2017recent}. After the improvements in the spatial resolution of the RS images, detecting objects in the images became more feasible, so researchers focused more on object-level segmentation and classification \cite{blaschke2001s, yan2006comparison}. Object-level classification aims to detect the objects and classify images depending on the relationships between these objects. This approach provides better results, however, it usually suffers from the diversity of object appearances in remote sensing scene images. More recently, especially after the increase in the availability of high spatial resolution aerial and satellite images, and public datasets with scene-based class labels, the research efforts are concentrated more on scene-level classification (RSSC), rather than pixel or object-level classification. 

RSSC is an image classification problem and computational solutions to this problem require the extraction of useful features from scene images. Xia et al. \cite{Xia_2017} categorized the feature extraction approaches in scene classification problem into three levels: low-level, mid-level and high level. Low-level and mid-level features are hand-crafted features that are coded manually using various computer vision algorithms \cite{inproceedingsSantos2010}, \cite{yang2008comparing}, \cite{chen2011evaluation}, \cite{luo2013indexing}. 

Despite the promising results presented in the literature, classifying a remote sensing image using hand-crafted features has its limitations in the RSSC domain. Representational power of features computed using traditional methods falls short against complex variations in remote sensing images. Instead, high-level features, which are also named deep features, are extracted using deep neural networks. Starting with the  seminal work of Krizhevsky et al. \cite{krizhevsky2012imagenet}, deep CNN models have been utilized successfully in various application domains and became one of the crucial methods, especially for image classification and semantic segmentation tasks from various domains \cite{SurveyClassification, SurveySegmentation}. Convolutional neural networks are deep learning structures where feature extraction and the classification phases are combined and learned end-to-end during training.  

Despite their end-to-end structures and high success rates in classification, CNN architectures have two major drawbacks: the complexity of training and the need for large amounts of labeled data. To avoid these two drawbacks, transfer learning methods are used to adapt pre-trained models, which are trained using ImageNet \cite{deng2009imagenet} dataset, to different domains.  ImageNet is a large natural image database that includes thousands of labeled examples on each class. Most of the well-known architectures, such as AlexNet\cite{krizhevsky2012imagenet}, DenseNet \cite{huang2018densely}, ResNet \cite{he2015deep}, are  trained using this dataset.  Retraining of these pre-trained networks with various approaches provides higher accuracies in other domains where labeled samples are scarce.

Recently, research in the RSSC domain has also been dominated by deep learning-based methods. Although there are large amounts of aerial or satellite images, there are only a few public benchmark datasets with labels in the RSSC domain \cite{marbhal2020evaluation}. To prevent the overfitting problem, which is a serious issue with deep model training using small datasets, custom-designed shallow architectures are utilized frequently, such as the one presented in Luus et al.'s work \cite{luus2015multiview}. They designed a small CNN with four convolutional layers. Nogueira et al, \cite{nogueira2017towards} compared the efficiency of training deep models from scratch and fine-tuning of pre-trained networks in the RSSC domain using different layers of CNN network as feature extractors.

The lack of a large number of labeled images in the RSSC domain and the advantages of transfer learning has led to the utilization of transfer learning-based approaches. Hu et al.\cite{hu2015transferring} and Cheng et al. \cite{cheng2016scene} analyzed how to use pre-trained CNNs in the RSSC domain.  Many follow-up works used different pre-trained CNN models and evaluated the performances of these models applying transfer learning and data augmentation techniques; in  \cite{castelluccio2015land} using GoogLeNet \cite{szegedy2014going} and CaffeNet \cite{jia2014caffe}, in \cite{scott2017training} using  CaffeNet, ResNet  and GoogLeNet models. In \cite{sen2019scene}, we evaluated the generalization performances of CNN models in the RSSC domain with a different perspective and reported that transfer learning of many moderate level CNN models performs better than their measured accuracies.

Recently, different methods have employed new strategies to increase the classification performances of CNN-based models on the available datasets. \cite{Cheng2018WhenDL} employed metric learning to address the inter-class similarity and intra-class diversity problems in RSSC domain; instead of the optimization based on cross-entropy loss, they optimized a discriminative  loss function during CNN trainings. In \cite{HierarchicalAtt&Fusion2020}, the authors used channel attention modules to improve the features that are obtained from different layers of the ResNet50 model. 
 In \cite{MetaLearning_2020_CVPR_Workshops}, model agnostic meta learning is applied in few shot classification setting for land use classification to handle high diversity of the scene images in RSSC domains. Some works utilized siamese and triplet networks \cite{liu2017scene, liu2019siamese}, some used model ensembles to increase the model performances \cite{zhang2015scene, scott2017fusion, scott2018enhanced}.

\begin{table*}[]
	\resizebox{\textwidth}{!}{%
		\begin{tabular}{lll}
			\hline
			\multicolumn{1}{c}{\textbf{\begin{tabular}[c]{@{}c@{}}First Level Classes\end{tabular}}} & \multicolumn{1}{c}{\textbf{Definition}}                                                                                                                           & \multicolumn{1}{c}{\textbf{Second Level Classes}}                                                                                                                                                 \\ \hline
			Buildings                                                                                   & \begin{tabular}[c]{@{}l@{}}Regions that \\ contain constructional activities\end{tabular}                                                                          & \begin{tabular}[c]{@{}l@{}}church, commercial areas, dense residential, \\ industrial area, medium residential, palace, storage tanks,\\ sparse residential, thermal power plants\end{tabular} \\ \hline
			Transportation                                                                              & \begin{tabular}[c]{@{}l@{}}Regions which are used by various\\  vehicles for transportation services\end{tabular}                                                 & \begin{tabular}[c]{@{}l@{}}airplane, airport, bridge, highway, runway, intersection, \\ mobile home park, overpass, parking lot, railway, \\ railway station, roundabout\end{tabular}          \\ \hline
			Lands Natural                                                                               & \begin{tabular}[c]{@{}l@{}}Regions that are not located in the urban area, \\ with no construction structures around or on them\end{tabular}                      & \begin{tabular}[c]{@{}l@{}}chapparal, circular farmlands, desert, forest, meadow, \\ mountain, rectangular farmland, terrace\end{tabular}                                                      \\ \hline
			Lands Constructed                                                                           & \begin{tabular}[c]{@{}l@{}}Regions, that are in the urban area. These are constructed structures \\ and generally buildings takes place around them.\end{tabular} & \begin{tabular}[c]{@{}l@{}}baseball court, basketball court, golf course, \\ stadium, tennis court\end{tabular}                                                                                \\ \hline
			Water Areas                                                                                 & Regions which predominantly contain  water or aquatic forms.                                                                                                      & \begin{tabular}[c]{@{}l@{}}beach, cloud, harbor, island, lake, river, sea-ice, \\ ships, snowberg wetlands\end{tabular}                                                                        \\ \hline
		\end{tabular}%
	}
	\caption{Two-Layered scheme of NW45 dataset. The \textit{First Level Classes} column refers the upper-level categories in the proposed hierarchical scheme. \textit{Second Level Classes} column contains the sub-categories under each first level category. }
	\label{tab:table1}
\end{table*}


Land use/cover changes are tracked to observe urban and rural development. Many countries have constructed schemes which are arranged in multiple levels \cite{LAGRO2005321}, \cite{CORINNE20190510}. Although classifying land use/cover classes in a multilevel hierarchically structured scheme is a common method, research using hierarchical, multi-layered, structures are fewer. 

Most of the works that use machine learning techniques create the hierarchical feature representation using clustering-based methods and then that hierarchical representation is used to improve the fine-grained predictions. In \cite{HierarchicalMetricLern_2017}, the authors organized the classes as binary trees using iterative max-margin clustering strategy and then trained each non-leaf node of this hierarchy to learn different metric transformations to increase the discrimination between the classes in that layer. In NW45, they reported $84.6\%$ classification accuracy with $80\%$ training data. Without metric learning training, the accuracy they report is $77.4\%$. (They use VGGNet-16 features in their work). In \cite{HierarhicalWassersteinCNN_2019}, the authors build a binary hierarchy three using the fusion of the deep features from AlexNet, GoogleNet, and VGGNet-16. They extracted features from these deep models and used clustering algorithms to determine categories in each branch of the hierarchy at that level. They obtained the best result in NW45, i.e. $96.37\%$, by fusing all the features from these three deep models while creating the binary hierarchy trees.

The hierarchical scheme that we apply in this work is different from the above-mentioned works. In the computer vision domain, there is a vast literature on hierarchic structures \cite{Survey2012}. One of the pre-deep learning era works that use object hierarchy trees in the classification of objects is \cite{ObjectHierarchy_ICCV_2007}. Relying on the fact that human cognition exploits hierarchic representation of world objects \cite{JOHNSON1998515}, they constructed sample object hierarchies as a tree structure and trained a classifier at each node of this tree using the natural images obtained from a set of objects from real life, such as car-side, school-bus under \textit{the closed frame vehicles category}; duck, ostrich under \textit{the winged animals' category}, etc. This is conceptually the most similar design to our approach in this domain. When identifying an object/scene, human beings follow a coarse to fine identification scheme. To exploit this; here, we create a two-layer hierarchy, where at each node we fine-tune a deep CNN model to the classes defined at that semantic layer using the ground truth labels for supervised training. Our work is also related to the hierarchical scheme proposed in \cite{ICCV2015_HDCNN} for object classification in ImageNet 1000-class dataset. The components in their model is also independently pretrained similar to our model. Based on our prior observations that the coarse-level classification is performed with high prediction probabilities, we avoided computationally expensive probabilistic averaging in the final prediction stage that is applied in \cite{ICCV2015_HDCNN}, which requires evaluation of all model predictions in the second level. Instead, we applied pruning to select the branch with the maximum prediction probability in the second level of our solution.  

Note that, the recent model performance improvement techniques, such as training model ensembles, metric learning, various forms of attention mechanisms, multimodel feature fusion, etc. are orthogonal to the research that we presented here. Since any performance improvement technique in the deep learning domain can be applied to the primary CNN models in our hierarchy, we are providing the performances of the vanilla models in our hierarchic scheme as a baseline hierarchic model. We obtained comparable classification performance, i.e. $94.10\pm0.65\%$ (see Table \ref{tab:table4}), with the existing state-of-the-art methods in this dataset, without applying any performance improvement techniques. 

\section{Materials}\label{sec:3}
We train our models using the NWPU-RESIS45 (NW45) dataset and two instances of the DenseNet architecture as the backbone CNN model. In this part, we will first briefly explain the NW45 dataset and provide the hierarchical scheme that we designed using the annotations in this dataset. Then, we provide the details of our hierarchical solution.

\subsection{NWPU-RESIS45 Dataset} \label{sec:24} 
NW45 is one of the largest datasets that is recently released by Cheng et al. \cite{Ref-NW45} for RSSC studies. It has 45 classes and each class includes 700 RGB images obtained from Google Earth images. The pixel size of all images is 256x256, and the spatial resolution varies from 20cm to 30m per pixel.  

This dataset contains a high degree of similarity between images from different classes and high diversity in images within the same class. These are the two important challenges that are faced frequently and need to be addressed in the RSSC domain.  Some sub-categories have remarkably high semantic overlaps, such as commercial area and industrial area, tennis court, and basketball court, which makes this data set more challenging compared to other smaller-scale datasets in this domain.  

\subsection{Hierarchical Interpretation of the NW45 Dataset}\label{sec:311} 
NW45 dataset is not originally designed and labeled in a multilayered hierarchical scheme. We re-organized the classes in a two-layer hierarchy and used that scheme to evaluate our model performances. To the best of our knowledge, this is the first work that tries this scheme on this dataset. In this context, we reinterpret the scenes in the original dataset in a two-layered scheme. In this interpretation, there are 5 main classes on the first layer: \textit{Buildings, Transportation, Natural Lands, Constructed Lands, and Water Areas}. All the 45 classes, which are originally defined in the NW45 dataset, are grouped semantically under these top-layer classes. The proposed two-layered hierarchical structure is summarized in Table \ref{tab:table1}.

\subsection{CNN Architectures }\label{sec:312} 

DenseNet architecture has different versions, depending on the number of convolutional layers, i.e. DenseNet-121, DenseNet-160, DenseNet-201 etc. Considering the performances, computational complexities, and the number of parameters in these models, we opted for an instance of DenseNet-121 that is pre-trained with the Imagenet \cite{deng2009imagenet} dataset. We experimented with two different model architectures as ablation studies in all the models that we developed in this research while forming the proposing hierarchical and non-hierarchical model structures.

\begin{table}[h]
\centering	
	\resizebox{0.4\textwidth}{!}{%

	\begin{tabular}{lrr}
		
		\hline
		\textbf{Parameter Type} & \multicolumn{1}{l}{\textbf{D121-Full}} & \multicolumn{1}{l}{\textbf{D121-Half}} \\ \hline
		\begin{tabular}[c]{@{}l@{}}Number of \\ convolutional layers\end{tabular} & 121       & 39        \\ \hline
		\begin{tabular}[c]{@{}l@{}}Number of\\ Total Weights\end{tabular}         & 7,042,629 & 1,603,269 \\ \hline
		\begin{tabular}[c]{@{}l@{}}Number of \\ Trained Weights\end{tabular}      & 6,958,981 & 1,587,973 \\ \hline
	\end{tabular}
	}
	\caption{Comparison of Full and Half model parameters. }
	\label{tab:table2}	
\end{table}

\textit{D121-Full}: This is the full form of the DenseNet-121 architecture. We changed only the number of classes in the last layer, i.e. softmax, of the model. We fine-tuned the last 7 layers of \textit{Conv4block[1...16]}. Other layers of the network are kept unchanged during training to protect the weights of the pre-trained model. Hence, in the hierarchic schemes, all the models, at the coarse and fine-levels, in the hierarchy tree share the corresponding fixed low-level layers of this model (Fig. \ref{fig:Fig_Flow}).

\textit{D121-Half}: This architecture is designed using only the first half of the DenseNet-121 model. We kept the layers only up to the \textit{pool3} layer and fine-tuned the last 4 layers of \textit{Conv3block[1..12]}. Hence, all the models in the hierarchy tree, at the coarse and fine-levels, share the corresponding fixed low-level layers of this model (Fig. \ref{fig:Fig_Flow}).

We used the compact network especially to observe the performance of using small models for high granularity categorical classification, (in the second level), in the proposed hierarchy. The comparison of these two CNN architectures according to their parameters are shown in Table-\ref{tab:table2}.

\begin{figure*}[ht!]
	
	\centering
	\includegraphics{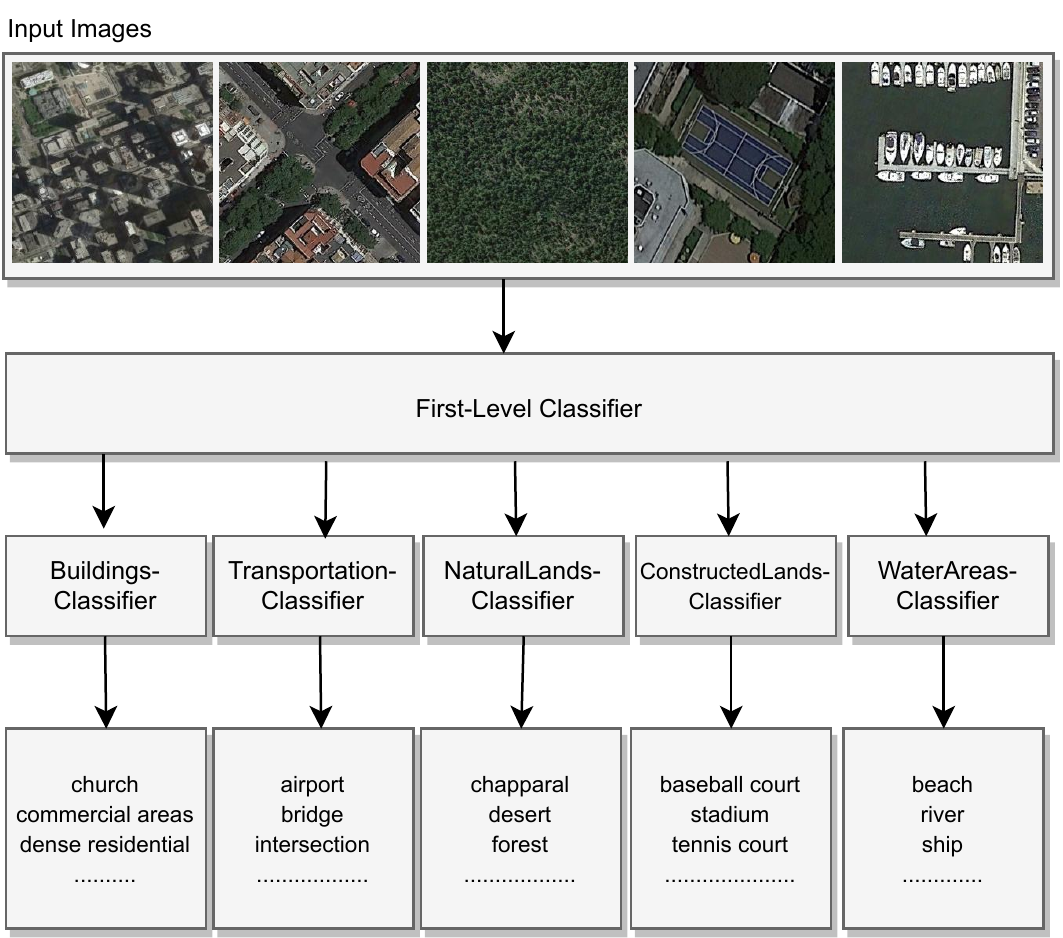}
	\caption{Structure of the proposed hierarchical model.} 
	\label{fig:Fig_ProposedModel}
\end{figure*}

\section{Proposed Model} \label{sec:32}
The semantic class relations in the proposed hierarchical model architecture are depicted in Fig.\ref{fig:Fig_ProposedModel}. (The flow diagram of the same model is depicted in Fig. \ref{fig:Fig_Flow}.) This architecture makes the classification process in two cascaded layers: in the first layer (the coarse-level), we have a \textit{First-level Classifier}. This model takes the input images and classifies them as one of the coarse-level categorical classes (Table-\ref{tab:table1}). According to the classification in the first layer, the input image is then sent to the related Second-level Classifier model, where the maximum predictive probability (MPP) is obtained as the coarse predictions, to get the finer-level class labels in that category. For example, if the input image is classified as \textit{Water Areas} using the First-level Classifier, with the MPP, then it is sent to WaterAreas Classifier to identify the fine-grained sub-categories of the related class, like river, beach, etc. A detailed explanation of the classifiers in our architectures is explained below:

\begin{itemize}
\item First-level Classifier: This is used for the classification of the image in a broader context, i.e. coarse-layer of the hierarchy. Output categories of the First-level Classifier are: \textit{Buildings, Transportation, Natural Lands, Constructed Lands, Water Areas} classes.

\item Second-level Classifiers: The classifiers after the First-level Classifier are referred to as the second-level classifiers, which are used to determine the fine-level categorical labels of given images. There are 5 second-level classifiers in our hierarchy; \textit{Buildings-Classifier, Transportation-Classifier, NaturalLands-Classifier, ConstructedLands-Classifier, WaterAreas-Classifier}. Each class in this level is trained using the related sub-level labels in the original NW45 dataset (Table \ref{tab:table1}).

\item Basic (Non-Hierarchic) Classifier: This classifier takes an input image and classifies it into one of the 45 categories that is defined in the NW45 dataset. This model is trained end-to-end using the ground truth labels of the NW45 dataset and detects the category of a given image directly.
\end{itemize}

\subsection{Model Training}\label{sec:training}

We trained all models using the 5-fold cross-validation method. We shuffled the data and created five-folds, each containing 448 sample images for training, 112 sample images for validation, and 140 sample images for testing. We applied data augmentation during training, by rotating the images 90, 180, and 270 degrees clockwise; and by taking the horizontal and vertical reflections of the images. To compare the performances of our complex and simple models, we trained both D121-Full and D121-Half model architectures as specified, for all the models in this work. 

\subsection{Experiment Settings}\label{sec:experiments}
 
We performed three different experiments using our 7 different models (Section \ref{sec:32}) which are trained using the same procedures explained in Section~\ref{sec:training}. For the hierarchical structure, we have a First-level Classifier, i.e. 5Class-Classifier, which has 5 categorical outputs, that are designed according to the proposed hierarchical scheme (Fig.~\ref{fig:Fig_ProposedModel}). For each output category, we have a Second-level Classifier, as we explained before. In the non-hierarchical scheme, we only have a Basic (non-hierarchic) Classifier, which classifies the images into 45-class categories that are defined in the NW45 dataset. We also refer to that model as a 45Class-Classifier. The details of the three experiments are explained below. 

\textbf{\textit{Experiment-1: Hierarchical top-down}}

After training all sub-models of the First-level and Second-level Classifier models, we constructed our proposed hierarchical structure (Fig.~\ref{fig:Fig_Flow}). In this experiment, First-level Classifier, which we also refer to as the 5Class-Classifier, takes an input image and classifies it as one of the 5 categorical outputs belonging to second-level categories. Then the input is classified with the related Second-level Classifier for which the MPP is obtained. 

\textbf{\textit{Experiment-2: Hierarchical top-down}}

In this setting, we skip the First-level classifier and directly send the input images to the related Second-level Classifier, which is determined using the ground-truth labels of the images. We want to see the test performance of the proposed hierarchical scheme when the First-level Classifier works with 100\% accuracy.

\textbf{\textit{Experiment-3: Hierarchical bottom-up}}

In this experiment, we perform a hierarchical classification from the bottom-up direction. We first classify a given image using our Basic (45Class) Classifier. According to the output of this classifier, the second-level category of the image is determined according to our hierarchical scheme (Table \ref{tab:table1}). Then, the input image is re-evaluated using the corresponding Second-level Classifier. For example, for a given \textit{river} image, assume that it is classified as a \textit{beach} by the Basic (45Class) Classifier. Since in our hierarchical scheme beach class is under the WaterAreas category, we send the same image to WaterAreas Classifier for re-evaluation. Then the category of the image is determined considering the sub-categories of the WaterAreas result, i.e. may be corrected as a river, in the second evaluation. The order is, therefore, bottom-up, where we follow the direction from a fine-category (bottom) to a coarse category (top). 

In addition to these experiments, we also evaluated the performance of the Basic (45Class) Classifier in a non-hierarchical setting.

\section{Results and Discussions} \label{sec:4}


\begin{table*}[]
\centering
	\resizebox{0.75\textwidth}{!}{%
		\begin{tabular}{llll}  
			\hline
			\multirow{2}{*}{ \textbf{Classifier Type} } & \multirow{2}{*}{\textbf{Classifier Name} } & \textbf{D121}  & \textbf{D121}   \\
			&                                            & \textbf{Half}  & \textbf{Full}   \\ 
			\hline
			First-level  Classifier                        & 5Class-Classifier                          & $96.54\pm1.62$
			& $97.17\pm0.39$          \\ 
			\hline
			Basic (Non Hierarchic) Classifier                        & 45Class-Classifier                         & $94.06\pm0.22$          & $95.40\pm0.66$           \\ 
			\hline
			&                                            &                &                 \\ 
			\hline
			Second-level Classifiers                  & Buildings-Classifier                       & $92.13\pm0.29$          & $93.44\pm1.09$           \\ 
			\hline
			Second-level Classifiers                  & Transportation-Classifier                  & $95.86\pm0.36$          & $96.68\pm0.29$           \\ 
			\hline
			Second-level Classifiers                  & NaturalLands-Classifier                    & $96.86 \pm0.54$          & $97.32\pm0.29$           \\ 
			\hline
			Second-level Classifiers                  & ConstructedLands-Classifier                & $97.38 \pm1.02$          & $98.69\pm0.39$           \\ 
			\hline
			Second-level Classifiers                 & WaterAreas-Classifier                      & $97.13\pm0.31$          & $98.11\pm0.33$           \\ 
			\hline
			&                                            &                &                
		\end{tabular}
	}
	
	\caption{Overall accuracy results (\%) of 7 classifiers. }
	\label{tab:table3}
\end{table*}

In this research, we want to analyze the performance  of the deep models that are configured to identify coarser semantic categories of images first, and then identify the finer category in the context of the coarse category in a hierarchical setting, in the challenging RSSC domain. We first start by evaluating the performances of the 7 basic models that we defined in detail in Section \ref{sec:experiments}. The experiment results are provided in Table~\ref{tab:table3}. As can be seen from the performances in this table, Second-level Classifiers perform very well on the related sub-category classification tasks. Except for the Building-Classifier, all the models perform higher than 96\% accuracy with the D121 Full model.  Considering the high performance of the First-level Classifier, i.e. more than 97\% accuracy, and the performances of the Second-level Classifiers, at a first glance, it seems as though the hierarchical structure will work better than the Basic (45Class) Classifier model. However, when we compute the overall accuracies, we observe that it's not actually the case (Table~\ref{tab:table4}).

Overall accuracies of the three experiments are provided in Table~\ref{tab:table4}, together with the performance of the non-hierarchic Basic Classifier. Our intuition behind modeling a hierarchical architecture was that the classification task in this model will be less challenging for the deep models in each layer. From the top-to-bottom, the model in the first level will only need to predict 5 broad categories, instead of learning all finer details of 45 categories; in the second level, each deep model will only need to learn the finer specifications of the less number of classes, which could potentially increase the overall accuracy rates.

 \begin{figure}[ht!]
	
	\centering
	\includegraphics[width=0.5\textwidth]{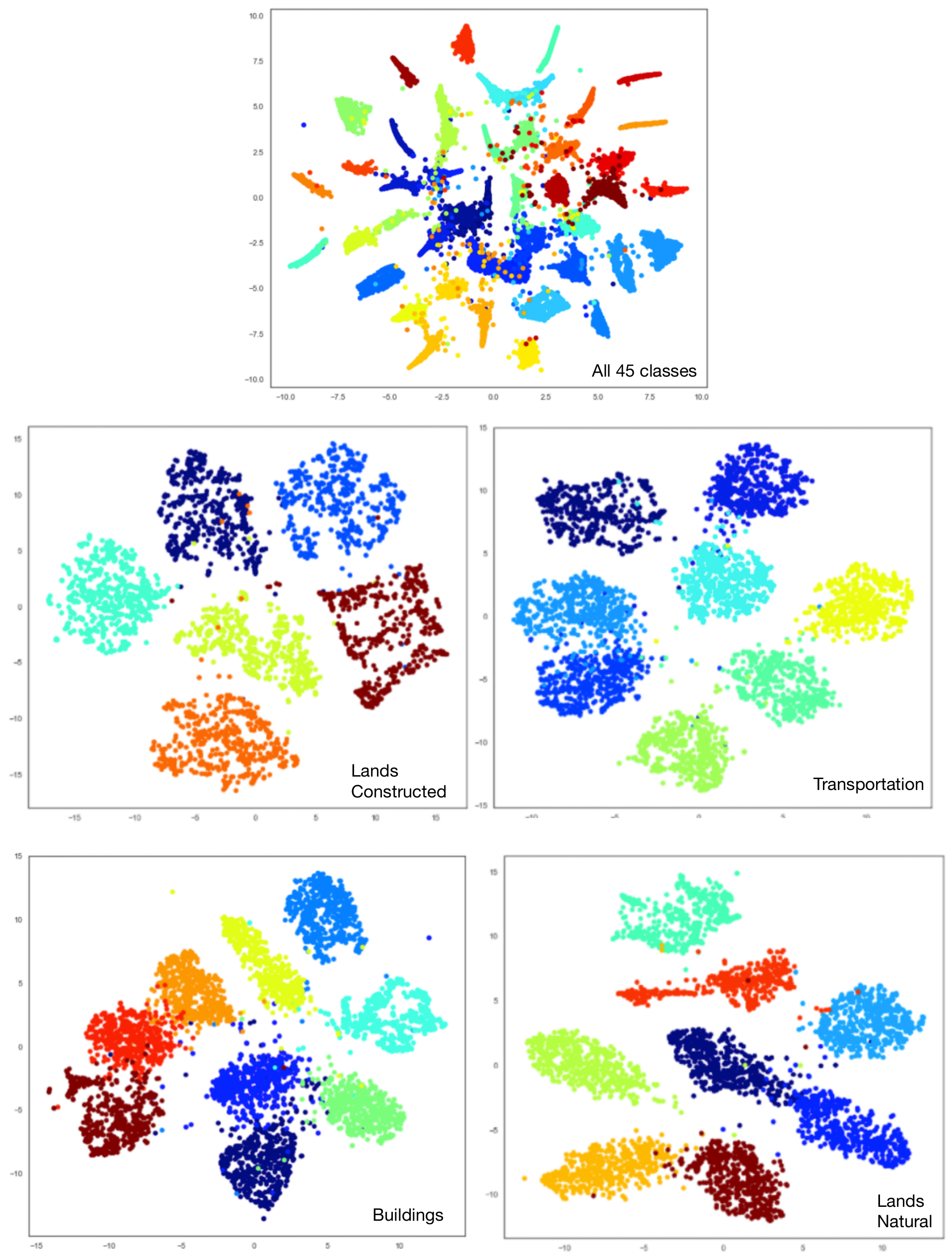}
	\caption{tSNE visualizations of feature space. Top row: non-hierarchic \textit{45Class Classifier}, middle, and bottom rows: hierarchic fine-level classifiers; from left to right \textit{Lands Constructed Classifier},  \textit{Transportation Classifier},  \textit{Buildings Classifier},  \textit{Lands Natural Classifier}.} 
	\label{fig:Fig_Projections}
\end{figure}

To understand the distribution of data in the trained models' embedding space better, we analyzed the feature distribution of all the models using the t-Distributed Stochastic Neighbor Embedding algorithm (tSNE) \cite{tSNEVis2008}. tSNE performs a nonlinear projection from a high dimensional data space to lower dimensions by optimizing a function to preserve neighborhood relation in the data as much as possible; hence very useful for visualizations of complex data distributions in higher dimensions. We took the feature vectors from the D121-Full model from the last layer, (right before the softmax layer, as an 1152 dimensional vector for each sample image) and project these vectors to 2-dimensional planes with tSNE. The results from 4 distinct models are displayed in Fig. \ref{fig:Fig_Projections}. As can be seen, 45 class non-hierarchic model's feature space is very crowded; although we can identify clusters of individual classes (each one is shown with a different color), some of the classes (in the central part of the distribution) have quite close proximities and many samples violating class boundaries between different clusters are visible. On the other hand, we can easily see that the hierarchic model's feature distributions are better-separated thanks to the dedicated capacity for a fewer number of classes. 

 \begin{figure}[ht!]
	
	\centering
	\includegraphics[width=0.5\textwidth]{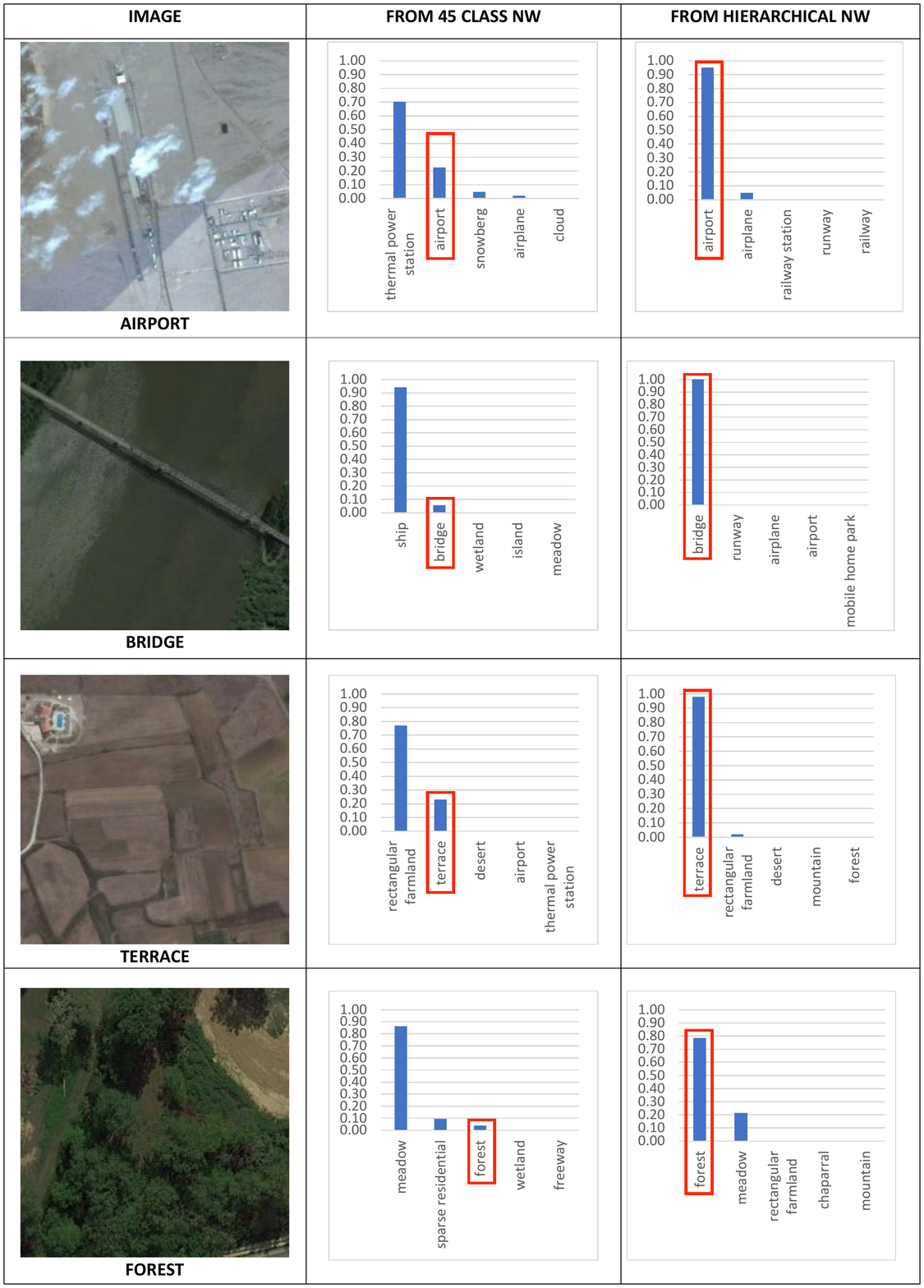}
	\caption{Predicted class probabilities. Correct class probabilities are depicted with red boxes. Left column: test image, middle column: non hierarchic (45Class) classifier outputs, right column: hierarchic (second-level) classifier outputs.} 
	\label{fig:Fig_OutputScores}
\end{figure}

As supported also by tSNE projections, independent evaluations from hierarchic levels show that our intuition is correct, however, the combined hierarchic model experiments - as we implemented here by pruning - showed us that it is not the case in the combined model. As can be seen, the performance of the primary Hierarchic model is less than all the other models, i.e. 92.61\% with the D121-Half and 94.10\% with the D121-Full model. In Experiment-2, the overall accuracy of the Second-level Classifiers, which is 96.73\% with the D121-Full model, is higher than that of the Basic Classifier performance, i.e. 95.40\%. We also analyzed the output prediction probabilities of Basic Classifier and Second-level classifiers for some samples that are misclassified by the Basic classifier. As shown in Fig. \ref{fig:Fig_OutputScores}, the confidence of the Second-level Classifiers are considerably higher for the true categories, while they are misclassified with a similar category with the Basic Classifier.

Our experiments show that the performance of the First-level Classifier in Experiment-1 affects the accuracy results considerably. Note that, in Experiment-2, we used the ground truth labels to determine the high-level category of each test sample. Even with the half-size model, Experiment-2 results are better than Experiment-1 accuracies. Experiment-3 performances show that re-evaluating the predictions of the Basic classifier does not improve the performance of the model as we thought at the beginning. The motivation was primarily to correct misclassifications in some upper categories that are caused by the inter-class similarities in the finer-level classes. The improvement is too small, hence in practice, it makes more sense to trust the predictions of Basic Classifier in one pass, rather than waiting for re-evaluation.   

\begin{table}
\resizebox{.5\textwidth}{!}{%
\begin{tabular}{lll} 
\hline
\multirow{2}{*}{\textbf{Experiment}}                                                                   & \textbf{D121}  & \textbf{D121}   \\
                                                                                                       & \textbf{Half}  & \textbf{Full}   \\ 
\hline
\begin{tabular}[c]{@{}l@{}}Experiment-1 (Hierarchic top-down)\\(5Class-Classifier + Second-Level Classifiers)\end{tabular}     & $92.61\pm0.20$          & $94.10\pm0.65$           \\ 
\hline
\begin{tabular}[c]{@{}l@{}}Experiment-2 (Hierarchic top-down)\\(Perfect 5Class-Classifier + Second-Level Classifiers)\end{tabular} & \textbf{$95.77\pm0.18$}          & \textbf{$96.73\pm0.36$}           \\ 
\hline
\begin{tabular}[c]{@{}l@{}}Experiment-3 (Hierarchic bottom-up)\\(45Class-Classifier + Second-Level Classifiers)\end{tabular}    & $94.07\pm0.22$          & $95.49\pm0.55$           \\ 

\hline
                                                                                                       &                &                 \\ 
\hline
\begin{tabular}[c]{@{}l@{}}Basic Classifier (Non Hierarchic)\\(45Class-Classifier)\end{tabular}    & $94.06\pm0.22$          & $95.40\pm0.66$           \\
\hline
\end{tabular}
}
\caption{Overall accuracy results (\%) of the experiments.}
\label{tab:table4}

\end{table}



\begin{table}
\resizebox{.5\textwidth}{!}{%
\begin{tabular}{l|lrr}
\multicolumn{1}{l}{\textbf{Model}}    & \textbf{Category}      & \multicolumn{1}{l}{\begin{tabular}[c]{@{}l@{}}\textbf{~Non~}\\\textbf{~Hierarchic~}\end{tabular}} & \multicolumn{1}{l}{\begin{tabular}[c]{@{}l@{}}\textbf{Hierarchic}\\\textbf{top-down}\end{tabular}}  \\ 
\hline
\multirow{6}{*}{\textbf{D121 - Half}} & Buildings               & $89.86 \pm 0.49$                                                                                                & $89.54 \pm 0.43$                                                                                               \\ 
\cline{2-4}
                                      & Constructed Lands       & $96.21 \pm 0.32 $                                                                                                & $95.17 \pm 0.85$                                                                                               \\ 
\cline{2-4}
                                      & Natural Lands           & $95.25 \pm 0.53$                                                                                                & $95.07 \pm 0.53$                                                                                               \\ 
\cline{2-4}
                                      & Transportation          & $93.94 \pm 0.43$                                                                                                & $93.36 \pm 0.57$                                                                                               \\ 
\cline{2-4}
                                      & Water Areas             & $95.73 \pm 0.46$                                                                                                & $90.97 \pm 0.42$                                                                                               \\ 
\cline{2-4}
                                      & \textbf{Overall Accuracy} & \textbf{$94.06 \pm 0.22$}                                                                                       & \textbf{$92.61 \pm 0.20$}                                                                                      \\ 
\hline
\multirow{6}{*}{\textbf{D121- Full}}  & Buildings               & $91.52 \pm 1.26$                                                                                                & $90.65 \pm 0.76$                                                                                               \\ 
\cline{2-4}
                                      & Constructed Lands       & $97.69\pm0.86$                                                                                                & $97.33\pm1.06$                                                                                               \\ 
\cline{2-4}
                                      & Natural Lands           & $96.16\pm0.85$                                                                                                & $95.96\pm0.47$                                                                                               \\ 
\cline{2-4}
                                      & Transportation          & $95.37\pm0.51$                                                                                                & $95.04\pm0.47$                                                                                               \\ 
\cline{2-4}
                                      & Water Areas             & $96.93\pm0.74$                                                                                                & $92.63\pm1.02$                                                                                               \\ 
\cline{2-4}
                                      & \textbf{Overall Accuracy} & \textbf{$95.40\pm0.66$}                                                                                       & \textbf{$94.10\pm0.65$}                                                                                      \\
\hline
\end{tabular}
}
	\caption{Empirical assessments (\% accuracies) of the Non Hierarchic  and Hierarchic (top-down) Classifiers. }
\label{tab:tabel5}
\end{table}
\begin{figure*}[ht!]
	\centering
	\includegraphics[scale=0.55]{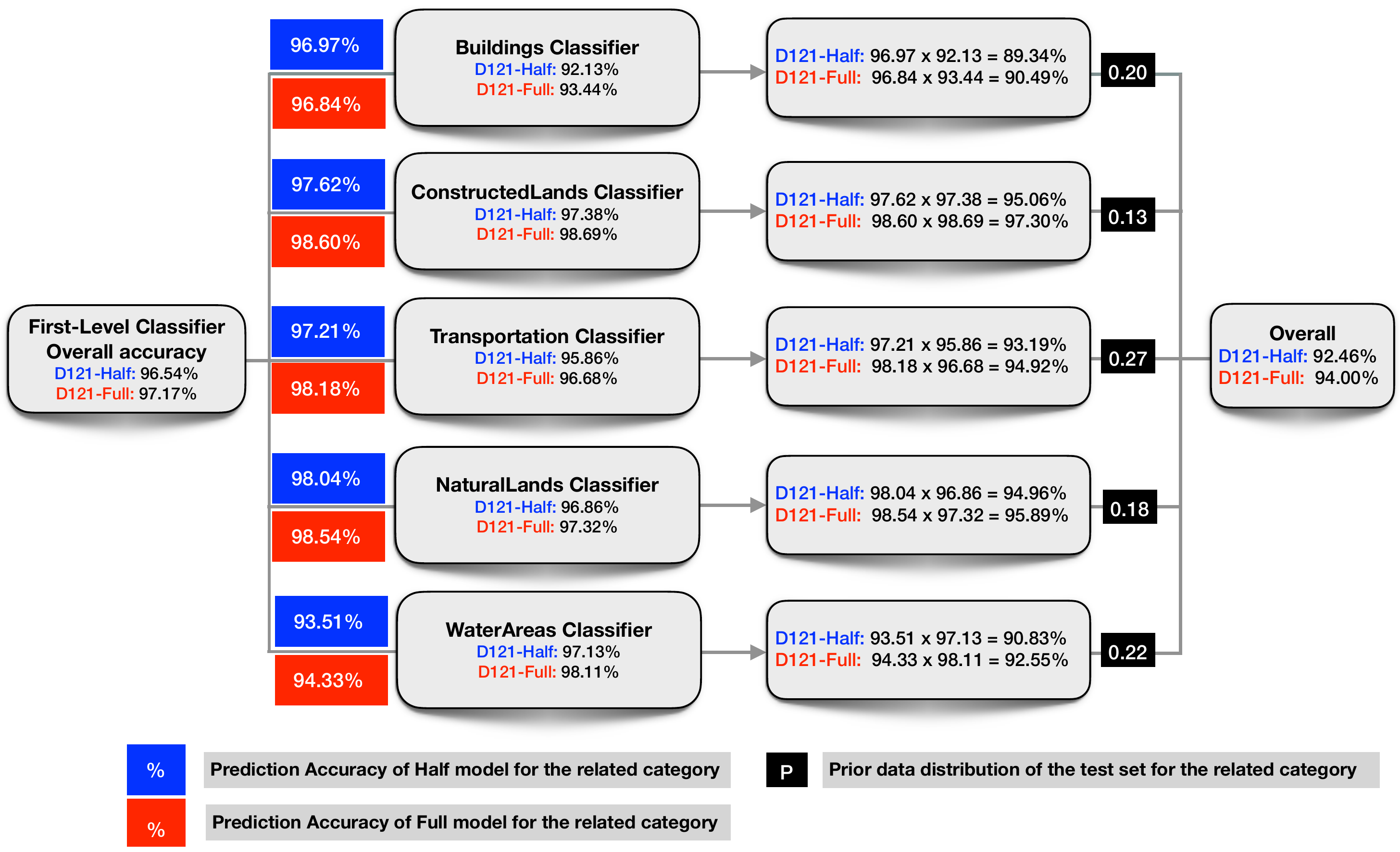}
	\caption{A statistical estimation of the classification accuracies. The accuracies are obtained by multiplication of the prediction accuracies in two layers. Accuracies in top (blue) boxes (on the left side of branches) show the categorical prediction accuracies for the Full First-level model; the bottom (red) boxes show that of the Half First-level model (for the category on the right). Overall prediction accuracies are computed using the prior test data distribution (depicted with the black boxes).}
	\label{fig:Fig_OverAllResults}
\end{figure*}

We also created the average accuracies of the Non-hierarchic Basic Classifier and Hierarchic top-down classifier (the model in Experiment-1) based on the 5 high-level categories that we defined in our hierarchical scheme (Table~\ref{tab:tabel5}). In both models, the fine level predictions are obtained first, i.e. river, bridge, etc. Then, the accuracies are computed on the related high-level category basis, considering the ground truth labels. For instance, for the N number of test samples in the Buildings category, we get the accuracy of the predictions of the models for the corresponding N samples. The empirical results provided in Table~\ref{tab:tabel5} show that the increase in the capacity of the model, i.e. using the D121-Full model, closes the performance gap between the Hierarchical model and the Basic model. The overall performance difference is 1.45\% with the D121 Half models, and 1.30\% with the D121-Full models. On the categorical basis, the hierarchic model makes most of its mistakes in the \textit{Water Areas} upper category.

All the results in our experiments, including the ones in Table~\ref{tab:tabel5}, are obtained using 5-fold cross-validation. In addition to the empirical analysis, we also conducted statistical analysis to validate the hierarchical model performance using the performances of the individual models in the hierarchy. The reported \textit{overall accuracies} that we computed in our experiments, in Table~\ref{tab:table3}, can be considered as the correct prediction probabilities of the models for each category. In our proposed hierarchical architecture, since each model evaluates the category of a given input independently from each other, the prediction probabilities of the hierarchical branches can be calculated using the multiplication of the probabilities of the cascaded layers. As a result, the overall performance of the model decreases in a two-layer scheme. The statistical analysis of our hierarchical model to estimate an approximate prediction accuracy for each upper category is depicted in Fig.~\ref{fig:Fig_OverAllResults}. As stated, each model's independent classification probabilities are obtained from Table~\ref{tab:table3} for this estimation. We also computed the overall prediction accuracies of the First-level model for each category; hence these accuracies are used as class conditional probabilities. Prior data distribution for the test set is used to weight the computed accuracies in each branch in the overall accuracy computation. As can be seen from the joint probabilities, although sub-category performances are slightly different, the overall accuracies of the statistical estimates are very close to the empirical results. This analysis validates our empirical results and better depicts the reason behind the performance reduction in our hierarchical models; although the individual model performances are superior in both levels, the small errors in two-cascaded layers are accumulated. For instance, the performance of the Buildings-Classifier in the hierarchical model (Second-level classifier), i.e. 93.44\%, is superior to that of the Basic Classifier, i.e. 91.52\%; yet, the joint probability with the First-level model takes it down, i.e. 90.65\%. As can also be seen from the statistical analysis, the Building class score, i.e. 90.49\% (Full model), is close to our empirical result and below that of the single-layer deep model, i.e. Basic Classifier.

The observation of the performance of Experiment-2 is also crucial (Table~\ref{tab:table4}). As also supported with the results in Table~\ref{tab:table3} and with the selected challenging samples in Fig. \ref{fig:Fig_OutputScores}, Second-level Classifiers perform very well, even with a small architecture like D121-Half.  They can produce accurate results when we want to classify a special region among only particular sub-categories. For example, if we want to find out forests in a particular region, we can use a Natural-Lands Classifier or we can consider training a more specific dichotomy for special queries when needed. 

\begin{table}[]
	\resizebox{0.5\textwidth}{!}{%
\begin{tabular}{|l|l|r|r|}
\hline
Model                      & Classifier                  & \multicolumn{1}{c|}{\begin{tabular}[c]{@{}c@{}}Total elapsed time\\ per batch (sec)\end{tabular}} & \multicolumn{1}{c|}{\begin{tabular}[c]{@{}c@{}}Mean time \\ per image (msec)\end{tabular}} \\ \hline
\multirow{2}{*}{D121-Full} & Basic Classifier (45 Class) & 4.44                                                                                              & 32.90                                                                                      \\ \cline{2-4} 
                           & Hierarchical Classifier     & 8.87                                                                                              & 65.71                                                                                      \\ \hline
\multirow{2}{*}{D121-Half} & Basic Classifier (45 Class) & 2.54                                                                                              & 18.86                                                                                      \\ \cline{2-4} 
                           & Hierarchical Classifier     & 5.09                                                                                              & 37.71                                                                                      \\ \hline
\end{tabular}
	}
	\caption{Average computational time estimations for a single batch of 135 images on a Tesla K80 GPU. }
	\label{tab:time6}
\end{table}


\begin{table}[t]
	\resizebox{0.5\textwidth}{!}{%
\begin{tabular}{|l|l|r|r|}
\hline
Model                      
& Classifier                  
& \multicolumn{1}{c|}{\begin{tabular}[c]{@{}c@{}}Total elapsed time\\ per batch (sec)\end{tabular}} & \multicolumn{1}{c|}{\begin{tabular}[c]{@{}c@{}}Mean time \\ per image (msec)\end{tabular}} \\ \hline
\multirow{2}{*}{D121-Full} & Basic Classifier (45 Class) & 10.36                                                                                              & 76.81                                                                                      \\ \cline{2-4} 
                           & Hierarchical Classifier     & 20.61                                                                                              & 152.68                                                                                      \\ \hline
\multirow{2}{*}{D121-Half} & Basic Classifier (45 Class) & 6.84                                                                                             & 50.67                                                                                      \\ \cline{2-4} 
                           & Hierarchical Classifier     & 13.75                                                                                              & 101.90                                                                                      \\ \hline
\end{tabular}
	}
	\caption{Average computational time estimations for a single batch of 135 images on Intel(R) Core(TM) i7-7700HQ 2.80GHz CPU. }
	\label{tab:time7}
\end{table}

\textbf{Computational Times of Models:} We estimated the computational times of the models in our hierarchical framework and the basic classifier and presented the estimated times in Tables \ref{tab:time6} and \ref{tab:time7}. The estimations are obtained both on GPU and CPU for a batch of 135 images multiple times (10 times) and the average of these estimations is reported. As can be seen, the running time of the hierarchical model is approximately 2 times more than the basic model (on both GPU and CPU runs) as we expected; since the hierarchic model contains the same CNN model with the Basic Classifier in a cascaded two-layer scheme. Note that, I/O times are not included in these estimations to be able to compare only the computational times of the deep models. 

\section{Conclusion} \label{sec:5}
In this work, we proposed a hierarchical framework to assess the performances of deep models in this scheme. All the models are trained using two instances of DenseNet 121 models, one with high capacity and one with low capacity. We obtained empirical performances for each model using a 5-fold cross-validation with the NW45 dataset. Our empirical analysis depicted that the hierarchical model performs comparable to the non-hierarchical model, yet it can not exceed the performance of our non-hierarchic model. We supported the estimated empirical results with the statistical estimates of the model prediction probabilities.

Image patches provided in the datasets in this domain do not always contain scenes related to one fine category; we have shown some examples in our previous study \cite{sen2019scene} depicting such ambiguities on NW45 and five other related datasets. Hence, there are some dataset biases imposed in the evaluations of the models in this domain. There is a need for standardization of the datasets in this domain and reducing the ambiguities in scene labeling. As of now, there is little agreement even on the category names in related datasets. In this work, we tried generating a two-layer hierarchical interpretation of one of the largest datasets in this field to contribute more to the semantic interpretation of the classes in a hierarchy and encourage future datasets to define similar schemes for further assessments.

\section*{Acknowledgements}
The numerical calculations reported in this paper were fully/partially performed at TUBITAK ULAKBIM, High Performance and Grid Computing Center (TRUBA resources).

\bibliographystyle{ieeetr}
\footnotesize \bibliography{RSSCv03}

\begin{thebibliography}{10}

\bibitem{Ref-NW45}
G.~Cheng, J.~Han, and X.~Lu, ``Remote sensing image scene classification:
  Benchmark and state of the art,'' {\em Proceedings of the IEEE}, vol.~105,
  no.~10, pp.~1865--1883, 2017.

\bibitem{phinn2012multi}
S.~R. Phinn, C.~M. Roelfsema, and P.~J. Mumby, ``Multi-scale, object-based
  image analysis for mapping geomorphic and ecological zones on coral reefs,''
  {\em International Journal of Remote Sensing}, vol.~33, no.~12,
  pp.~3768--3797, 2012.

\bibitem{li2013object}
X.~Li and G.~Shao, ``Object-based urban vegetation mapping with high-resolution
  aerial photography as a single data source,'' {\em International journal of
  remote sensing}, vol.~34, no.~3, pp.~771--789, 2013.

\bibitem{mishra2014mapping}
N.~B. Mishra and K.~A. Crews, ``Mapping vegetation morphology types in a dry
  savanna ecosystem: Integrating hierarchical object-based image analysis with
  random forest,'' {\em International Journal of Remote Sensing}, vol.~35,
  no.~3, pp.~1175--1198, 2014.

\bibitem{kim2009forest}
M.~Kim, M.~Madden, and T.~A. Warner, ``Forest type mapping using
  object-specific texture measures from multispectral ikonos imagery,'' {\em
  Photogrammetric Engineering \& Remote Sensing}, vol.~75, no.~7, pp.~819--829,
  2009.

\bibitem{janssen1992knowledge}
L.~L. Janssen and H.~Middelkoop, ``Knowledge-based crop classification of a
  landsat thematic mapper image,'' {\em International Journal of Remote
  Sensing}, vol.~13, no.~15, pp.~2827--2837, 1992.

\bibitem{stumpf2011object}
A.~Stumpf and N.~Kerle, ``Object-oriented mapping of landslides using random
  forests,'' {\em Remote sensing of environment}, vol.~115, no.~10,
  pp.~2564--2577, 2011.

\bibitem{cheng2013automatic}
G.~Cheng, L.~Guo, T.~Zhao, J.~Han, H.~Li, and J.~Fang, ``Automatic landslide
  detection from remote-sensing imagery using a scene classification method
  based on bovw and plsa,'' {\em International Journal of Remote Sensing},
  vol.~34, no.~1, pp.~45--59, 2013.

\bibitem{chen2014pyramid}
S.~Chen and Y.~Tian, ``Pyramid of spatial relatons for scene-level land use
  classification,'' {\em IEEE Transactions on Geoscience and Remote Sensing},
  vol.~53, no.~4, pp.~1947--1957, 2014.

\bibitem{chen2016land}
C.~Chen, B.~Zhang, H.~Su, W.~Li, and L.~Wang, ``Land-use scene classification
  using multi-scale completed local binary patterns,'' {\em Signal, image and
  video processing}, vol.~10, no.~4, pp.~745--752, 2016.

\bibitem{Cheng_2020}
G.~Cheng, X.~Xie, J.~Han, L.~Guo, and G.-S. Xia, ``Remote sensing image scene
  classification meets deep learning: Challenges, methods, benchmarks, and
  opportunities,'' {\em IEEE Journal of Selected Topics in Applied Earth
  Observations and Remote Sensing}, vol.~13, p.~3735–3756, 2020.

\bibitem{blaschke2001s}
T.~Blaschke and J.~Strobl, ``What’s wrong with pixels? some recent
  developments interfacing remote sensing and gis,'' {\em Zeitschrift f{\"u}r
  Geoinformationssysteme}, pp.~12--17, 2001.

\bibitem{blaschke2014geographic}
T.~Blaschke, G.~J. Hay, M.~Kelly, S.~Lang, P.~Hofmann, E.~Addink, R.~Q.
  Feitosa, F.~Van~der Meer, H.~Van~der Werff, F.~Van~Coillie, {\em et~al.},
  ``Geographic object-based image analysis--towards a new paradigm,'' {\em
  ISPRS journal of photogrammetry and remote sensing}, vol.~87, pp.~180--191,
  2014.

\bibitem{druaguct2006automated}
L.~Dr{\u{a}}gu{\c{t}} and T.~Blaschke, ``Automated classification of landform
  elements using object-based image analysis,'' {\em Geomorphology}, vol.~81,
  no.~3-4, pp.~330--344, 2006.

\bibitem{eisank2011generic}
C.~Eisank, L.~Dr{\u{a}}gu{\c{t}}, and T.~Blaschke, ``A generic procedure for
  semantics-oriented landform classification using object-based image
  analysis,'' {\em Geomorphometry}, vol.~2011, pp.~125--128, 2011.

\bibitem{bratasanu2010bridging}
D.~Bratasanu, I.~Nedelcu, and M.~Datcu, ``Bridging the semantic gap for
  satellite image annotation and automatic mapping applications,'' {\em IEEE
  Journal of Selected Topics in Applied Earth Observations and Remote Sensing},
  vol.~4, no.~1, pp.~193--204, 2010.

\bibitem{zhao2016feature}
L.~Zhao, P.~Tang, and L.~Huo, ``Feature significance-based
  multibag-of-visual-words model for remote sensing image scene
  classification,'' {\em Journal of Applied Remote Sensing}, vol.~10, no.~3,
  p.~035004, 2016.

\bibitem{xia2015accurate}
G.-S. Xia, Z.~Wang, C.~Xiong, and L.~Zhang, ``Accurate annotation of remote
  sensing images via active spectral clustering with little expert knowledge,''
  {\em Remote Sensing}, vol.~7, no.~11, pp.~15014--15045, 2015.

\bibitem{cheng2015effective}
G.~Cheng, J.~Han, L.~Guo, Z.~Liu, S.~Bu, and J.~Ren, ``Effective and efficient
  midlevel visual elements-oriented land-use classification using vhr remote
  sensing images,'' {\em IEEE Transactions on Geoscience and Remote Sensing},
  vol.~53, no.~8, pp.~4238--4249, 2015.

\bibitem{zhang2013semi}
Y.~Zhang, X.~Zheng, G.~Liu, X.~Sun, H.~Wang, and K.~Fu, ``Semi-supervised
  manifold learning based multigraph fusion for high-resolution remote sensing
  image classification,'' {\em IEEE Geoscience and Remote Sensing Letters},
  vol.~11, no.~2, pp.~464--468, 2013.

\bibitem{zheng2012automatic}
X.~Zheng, X.~Sun, K.~Fu, and H.~Wang, ``Automatic annotation of satellite
  images via multifeature joint sparse coding with spatial relation
  constraint,'' {\em IEEE Geoscience and Remote Sensing Letters}, vol.~10,
  no.~4, pp.~652--656, 2012.

\bibitem{Ref-UCM}
Y.~Yang and S.~Newsam, ``Bag-of-visual-words and spatial extensions for
  land-use classification,'' in {\em Proceedings of the 18th SIGSPATIAL
  international conference on advances in geographic information systems},
  pp.~270--279, 2010.

\bibitem{zhong2015scene}
Y.~Zhong, Q.~Zhu, and L.~Zhang, ``Scene classification based on the
  multifeature fusion probabilistic topic model for high spatial resolution
  remote sensing imagery,'' {\em IEEE Transactions on Geoscience and Remote
  Sensing}, vol.~53, no.~11, pp.~6207--6222, 2015.

\bibitem{zhang2016semantic}
J.~Zhang, T.~Li, X.~Lu, and Z.~Cheng, ``Semantic classification of
  high-resolution remote-sensing images based on mid-level features,'' {\em
  IEEE Journal of Selected Topics in Applied Earth Observations and Remote
  Sensing}, vol.~9, no.~6, pp.~2343--2353, 2016.

\bibitem{hu2015unsupervised}
F.~Hu, G.-S. Xia, Z.~Wang, X.~Huang, L.~Zhang, and H.~Sun, ``Unsupervised
  feature learning via spectral clustering of multidimensional patches for
  remotely sensed scene classification,'' {\em IEEE Journal of Selected Topics
  in Applied Earth Observations and Remote Sensing}, vol.~8, no.~5, 2015.

\bibitem{Ref-AID}
G.~Xia, J.~Hu, F.~Hu, B.~Shi, X.~Bai, Y.~Zhong, and L.~Zhang, ``{AID:} {A}
  benchmark dataset for performance evaluation of aerial scene
  classification,'' {\em CoRR}, vol.~abs/1608.05167, 2016.

\bibitem{Ref-WHU}
D.~Dai and W.~Yang, ``Satellite image classification via two-layer sparse
  coding with biased image representation,'' {\em IEEE Geosci. Remote Sensing
  Lett.}, vol.~8, pp.~173--176, 01 2011.

\bibitem{Ref-RSICB}
H.~Li, C.~Tao, Z.~Wu, J.~Chen, J.~Gong, and M.~Deng, ``{RSI-CB:} {A} large
  scale remote sensing image classification benchmark via crowdsource data,''
  {\em CoRR}, vol.~abs/1705.10450, 2017.

\bibitem{deng2009imagenet}
J.~Deng, W.~Dong, R.~Socher, L.-J. Li, K.~Li, and L.~Fei-Fei, ``Imagenet: A
  large-scale hierarchical image database,'' in {\em 2009 IEEE conference on
  computer vision and pattern recognition}, pp.~248--255, Ieee, 2009.

\bibitem{huang2018densely}
G.~Huang, Z.~Liu, L.~van~der Maaten, and K.~Q. Weinberger, ``Densely connected
  convolutional networks,'' 2018.

\bibitem{ICCV2015_HDCNN}
Z.~Yan, H.~Zhang, R.~Piramuthu, V.~Jagadeesh, D.~DeCoste, W.~Di, and Y.~Yu,
  ``Hd-cnn: Hierarchical deep convolutional neural networks for large scale
  visual recognition,'' {\em 2015 IEEE International Conference on Computer
  Vision (ICCV)}, pp.~2740--2748, 2015.

\bibitem{sen2019scene}
O.~Sen and H.~Y. Keles, ``Scene recognition with deep learning methods using
  aerial images,'' in {\em 2019 27th Signal Processing and Communications
  Applications Conference (SIU)}, pp.~1--4, IEEE, 2019.

\bibitem{tuia2011survey}
D.~Tuia, M.~Volpi, L.~Copa, M.~Kanevski, and J.~Munoz-Mari, ``A survey of
  active learning algorithms for supervised remote sensing image
  classification,'' {\em IEEE Journal of Selected Topics in Signal Processing},
  vol.~5, no.~3, pp.~606--617, 2011.

\bibitem{he2017recent}
L.~He, J.~Li, C.~Liu, and S.~Li, ``Recent advances on spectral--spatial
  hyperspectral image classification: An overview and new guidelines,'' {\em
  IEEE Transactions on Geoscience and Remote Sensing}, vol.~56, no.~3,
  pp.~1579--1597, 2017.

\bibitem{yan2006comparison}
G.~Yan, J.-F. Mas, B.~Maathuis, Z.~Xiangmin, and P.~Van~Dijk, ``Comparison of
  pixel-based and object-oriented image classification approaches—a case
  study in a coal fire area, wuda, inner mongolia, china,'' {\em International
  Journal of Remote Sensing}, vol.~27, no.~18, pp.~4039--4055, 2006.

\bibitem{Xia_2017}
G.-S. Xia, J.~Hu, F.~Hu, B.~Shi, X.~Bai, Y.~Zhong, L.~Zhang, and X.~Lu, ``Aid:
  A benchmark data set for performance evaluation of aerial scene
  classification,'' {\em IEEE Transactions on Geoscience and Remote Sensing},
  vol.~55, p.~3965–3981, Jul 2017.

\bibitem{inproceedingsSantos2010}
J.~dos Santos, O.~Penatti, and R.~Torres, ``Evaluating the potential of texture
  and color descriptors for remote sensing image retrieval and
  classification.,'' vol.~2, pp.~203--208, 01 2010.

\bibitem{yang2008comparing}
Y.~Yang and S.~Newsam, ``Comparing sift descriptors and gabor texture features
  for classification of remote sensed imagery,'' in {\em 2008 15th IEEE
  international conference on image processing}, pp.~1852--1855, IEEE, 2008.

\bibitem{chen2011evaluation}
L.~Chen, W.~Yang, K.~Xu, and T.~Xu, ``Evaluation of local features for scene
  classification using vhr satellite images,'' in {\em 2011 Joint Urban Remote
  Sensing Event}, pp.~385--388, IEEE, 2011.

\bibitem{luo2013indexing}
B.~Luo, S.~Jiang, and L.~Zhang, ``Indexing of remote sensing images with
  different resolutions by multiple features,'' {\em IEEE Journal of Selected
  Topics in Applied Earth Observations and Remote Sensing}, vol.~6, no.~4,
  pp.~1899--1912, 2013.

\bibitem{krizhevsky2012imagenet}
A.~Krizhevsky, I.~Sutskever, and G.~E. Hinton, ``Imagenet classification with
  deep convolutional neural networks,'' {\em Advances in neural information
  processing systems}, vol.~25, pp.~1097--1105, 2012.

\bibitem{SurveyClassification}
M.~Sornam, K.~Muthusubash, and V.~Vanitha, ``A survey on image classification
  and activity recognition using deep convolutional neural network
  architecture,'' in {\em 2017 Ninth International Conference on Advanced
  Computing (ICoAC)}, pp.~121--126, 2017.

\bibitem{SurveySegmentation}
F.~Cao and Q.~Bao, ``A survey on image semantic segmentation methods with
  convolutional neural network,'' in {\em 2020 International Conference on
  Communications, Information System and Computer Engineering (CISCE)},
  pp.~458--462, 2020.

\bibitem{he2015deep}
K.~He, X.~Zhang, S.~Ren, and J.~Sun, ``Deep residual learning for image
  recognition,'' 2015.

\bibitem{marbhal2020evaluation}
S.~Marbhal and M.~Kumar, ``Evaluation of datasets for cnn based image
  classification,'' 2020.

\bibitem{luus2015multiview}
F.~P. Luus, B.~P. Salmon, F.~Van~den Bergh, and B.~T.~J. Maharaj, ``Multiview
  deep learning for land-use classification,'' {\em IEEE Geoscience and Remote
  Sensing Letters}, vol.~12, no.~12, pp.~2448--2452, 2015.

\bibitem{nogueira2017towards}
K.~Nogueira, O.~A. Penatti, and J.~A. Dos~Santos, ``Towards better exploiting
  convolutional neural networks for remote sensing scene classification,'' {\em
  Pattern Recognition}, vol.~61, pp.~539--556, 2017.

\bibitem{hu2015transferring}
F.~Hu, G.-S. Xia, J.~Hu, and L.~Zhang, ``Transferring deep convolutional neural
  networks for the scene classification of high-resolution remote sensing
  imagery,'' {\em Remote Sensing}, vol.~7, no.~11, pp.~14680--14707, 2015.

\bibitem{cheng2016scene}
G.~Cheng, C.~Ma, P.~Zhou, X.~Yao, and J.~Han, ``Scene classification of high
  resolution remote sensing images using convolutional neural networks,'' in
  {\em 2016 IEEE International Geoscience and Remote Sensing Symposium
  (IGARSS)}, pp.~767--770, IEEE, 2016.

\bibitem{castelluccio2015land}
M.~Castelluccio, G.~Poggi, C.~Sansone, and L.~Verdoliva, ``Land use
  classification in remote sensing images by convolutional neural networks,''
  {\em arXiv preprint arXiv:1508.00092}, 2015.

\bibitem{szegedy2014going}
C.~Szegedy, W.~Liu, Y.~Jia, P.~Sermanet, S.~Reed, D.~Anguelov, D.~Erhan,
  V.~Vanhoucke, and A.~Rabinovich, ``Going deeper with convolutions,'' 2014.

\bibitem{jia2014caffe}
Y.~Jia, E.~Shelhamer, J.~Donahue, S.~Karayev, J.~Long, R.~Girshick,
  S.~Guadarrama, and T.~Darrell, ``Caffe: Convolutional architecture for fast
  feature embedding,'' 2014.

\bibitem{scott2017training}
G.~J. Scott, M.~R. England, W.~A. Starms, R.~A. Marcum, and C.~H. Davis,
  ``Training deep convolutional neural networks for land--cover classification
  of high-resolution imagery,'' {\em IEEE Geoscience and Remote Sensing
  Letters}, vol.~14, no.~4, pp.~549--553, 2017.

\bibitem{Cheng2018WhenDL}
G.~Cheng, C.~Yang, X.~Yao, L.~Guo, and J.~Han, ``When deep learning meets
  metric learning: Remote sensing image scene classification via learning
  discriminative cnns,'' {\em IEEE Transactions on Geoscience and Remote
  Sensing}, vol.~56, pp.~2811--2821, 2018.

\bibitem{HierarchicalAtt&Fusion2020}
D.~Yu, H.~Guo, Q.~Xu, J.~Lu, C.~Zhao, and Y.~Lin, ``Hierarchical attention and
  bilinear fusion for remote sensing image scene classification,'' {\em IEEE
  Journal of Selected Topics in Applied Earth Observations and Remote Sensing},
  vol.~13, pp.~6372--6383, 2020.

\bibitem{MetaLearning_2020_CVPR_Workshops}
M.~Russwurm, S.~Wang, M.~Korner, and D.~Lobell, ``Meta-learning for few-shot
  land cover classification,'' in {\em Proceedings of the IEEE/CVF Conference
  on Computer Vision and Pattern Recognition (CVPR) Workshops}, June 2020.

\bibitem{liu2017scene}
Y.~Liu and C.~Huang, ``Scene classification via triplet networks,'' {\em IEEE
  Journal of Selected Topics in Applied Earth Observations and Remote Sensing},
  vol.~11, no.~1, pp.~220--237, 2017.

\bibitem{liu2019siamese}
X.~Liu, Y.~Zhou, J.~Zhao, R.~Yao, B.~Liu, and Y.~Zheng, ``Siamese convolutional
  neural networks for remote sensing scene classification,'' {\em IEEE
  Geoscience and Remote Sensing Letters}, vol.~16, no.~8, pp.~1200--1204, 2019.

\bibitem{zhang2015scene}
F.~Zhang, B.~Du, and L.~Zhang, ``Scene classification via a gradient boosting
  random convolutional network framework,'' {\em IEEE Transactions on
  Geoscience and Remote Sensing}, vol.~54, no.~3, pp.~1793--1802, 2015.

\bibitem{scott2017fusion}
G.~J. Scott, R.~A. Marcum, C.~H. Davis, and T.~W. Nivin, ``Fusion of deep
  convolutional neural networks for land cover classification of
  high-resolution imagery,'' {\em IEEE Geoscience and Remote Sensing Letters},
  vol.~14, no.~9, pp.~1638--1642, 2017.

\bibitem{scott2018enhanced}
G.~J. Scott, K.~C. Hagan, R.~A. Marcum, J.~A. Hurt, D.~T. Anderson, and C.~H.
  Davis, ``Enhanced fusion of deep neural networks for classification of
  benchmark high-resolution image data sets,'' {\em IEEE Geoscience and Remote
  Sensing Letters}, vol.~15, no.~9, pp.~1451--1455, 2018.

\bibitem{LAGRO2005321}
J.~LaGro, ``Land-use classification,'' in {\em Encyclopedia of Soils in the
  Environment} (D.~Hillel, ed.), pp.~321 -- 328, Oxford: Elsevier, 2005.

\bibitem{CORINNE20190510}
K.~Barbara, B.~György, H.~Gerard, and A.~Stephan, ``Updated clc illustrated
  nomenclature guidelines,'' in {\em Updated CLC illustrated nomenclature
  guidelines}, European Environment Agency, 2019.

\bibitem{HierarchicalMetricLern_2017}
A.~Goel and B.~Banerjee, ``Hierarchical metric learning for fine grained image
  classification,'' {\em CoRR}, vol.~abs/1708.01494, 2017.

\bibitem{HierarhicalWassersteinCNN_2019}
Y.~Liu, C.~Y. Suen, Y.~Liu, and L.~Ding, ``Scene classification using
  hierarchical wasserstein cnn,'' {\em IEEE Transactions on Geoscience and
  Remote Sensing}, vol.~57, no.~5, pp.~2494--2509, 2019.

\bibitem{Survey2012}
A.-M. Tousch, S.~Herbin, and J.-Y. Audibert, ``Semantic hierarchies for image
  annotation: A survey,'' {\em Pattern Recogn.}, vol.~45, p.~333–345, Jan.
  2012.

\bibitem{ObjectHierarchy_ICCV_2007}
A.~Zweig and D.~Weinshall, ``Exploiting object hierarchy: Combining models from
  different category levels,'' in {\em 2007 IEEE 11th International Conference
  on Computer Vision}, pp.~1--8, 2007.

\bibitem{JOHNSON1998515}
K.~E. Johnson and A.~T. Eilers, ``Effects of knowledge and development on
  subordinate level categorization,'' {\em Cognitive Development}, vol.~13,
  no.~4, pp.~515--545, 1998.

\bibitem{tSNEVis2008}
L.~van~der Maaten and G.~Hinton, ``Visualizing high-dimensional data using
  t-sne,'' 2008.

\end{thebibliography}

\end{document}